\documentclass[10pt,letterpaper]{article}

\usepackage{hyperref} 
\usepackage{cogsci}

\cogscifinalcopy 

\usepackage{pslatex}
\usepackage{apacite}
\usepackage{float} 
                   
\usepackage{courier} 
\usepackage{csquotes} 
\usepackage{setspace} 
\usepackage{footnote} 
\usepackage{amsmath} 
\usepackage{graphicx} 
\usepackage{booktabs} 
\usepackage{cuted} 
\usepackage{etoolbox} 

\title{Compositional Processing Emerges in Neural Networks Solving Math Problems}

\author{{\large \bf Jacob Russin$^{1}$ (jlrussin@ucdavis.edu)} \\
  \And {\large \bf Roland Fernandez$^{2}$ (rfernand@microsoft.com)} \\
  \AND {\large \bf Hamid Palangi$^2$ (hpalangi@microsoft.com)} \\
  \And {\large \bf Eric Rosen$^{3}$ (erosen27@jhu.edu)} \\
  \AND {\large \bf Nebojsa Jojic$^{2}$ (jojic@microsoft.com)} \\
  \And {\large \bf Paul Smolensky$^{2,3}$ (psmo@microsoft.com)} \\
  \AND {\large \bf Jianfeng Gao$^{2}$ (jfgao@microsoft.com)} \\
  }

\begin{document}

\maketitle

\begin{strip}
{\centering
    $^1$ Department of Psychology, UC Davis; $^2$ Microsoft Research, Redmond;  
    
    $^3$ Department of Cognitive Science, Johns Hopkins University
\par}
\end{strip}

\begin{abstract}
A longstanding question in cognitive science concerns the learning mechanisms underlying compositionality in human cognition.
Humans can infer the structured relationships (e.g., grammatical rules) implicit in their sensory observations (e.g., auditory speech), and use this knowledge to guide the composition of simpler meanings into complex wholes.
Recent progress in artificial neural networks has shown that when large models are trained on enough linguistic data, grammatical structure emerges in their representations.
We extend this work to the domain of mathematical reasoning, where it is possible to formulate precise hypotheses about how meanings (e.g., the quantities corresponding to numerals) should be composed according to structured rules (e.g., order of operations).  
Our work shows that neural networks are not only able to infer something about the structured relationships implicit in their training data, but can also deploy this knowledge to guide the composition of individual meanings into composite wholes.

\end{abstract}

\section{Introduction}
A key to the power of human cognition is the principle of compositionality \cite{hinzen2012oxford}: complex stimuli such as sentences are understood by 1) recognizing their relevant subcomponents, 2) extracting the meanings of these subcomponents, and 3) combining these partial meanings according to structured rules to produce a final result --- the meaning of the whole stimulus \cite{MartinBaggio20a}.
This allows a potentially infinite number of stimuli to be understood as novel compositions of familiar parts. 
Compositionality was explicitly built into traditional symbolic artificial intelligence (AI) systems, but in the currently dominant approach --- deep neural networks --- encodings are continuous vectors that do not overtly possess the relevant kind of compositionality \cite{fodor1988connectionism, LakeBaroni18, LakeUllmanTenenbaumEtAl17, McClellandHillRudolphEtAl20a}.

Although these recent models are the most powerful AI systems ever created, explaining how they achieve their remarkable success is a profound mystery --- a grand challenge for current science.
Is it possible that deep neural networks are somehow implicitly exploiting the principle of compositionality (See Fig.~\ref{fig:fig1}A)?

State-of-the-art neural networks called transformers \cite{VaswaniShazeerParmarEtAl17b} have shown impressive performance on many natural language tasks \cite{DevlinChangLeeEtAl19b}. 
Briefly, these networks learn to encode each symbol in their inputs with a high-dimensional vector that is successively refined over several layers by adding information from other symbols in the input. 
There is evidence that these networks reflect in their activation patterns the relevant linguistic subcomponents in their inputs \cite{LinzenBaroni21, ManningClarkHewittEtAl20a, TenneyDasPavlick19a}.
However, it is unclear how, if at all, these models use such structure to extract and compose the meanings of parts to succeed at their tasks \cite{LakeUllmanTenenbaumEtAl17}. 

Studying this question in the natural language setting is challenging because it is extremely difficult to precisely characterize the meaning of a part of a natural language expression.
We therefore study the question in a domain where the meaning of a subcomponent is clear: arithmetic expressions.
The `meaning' of a part --- a sub-expression --- is simply its numerical value, and the principle of compositionality holds perfectly: the value of $4*5 + 2*3$ is precisely obtained by taking the meanings of the sub-expressions $4*5$ and $2*3$ (i.e., the quantities denoted by the numerals `$20$' and `$6$') and composing them with the addition operation to derive the meaning of the entire expression, the number written `$26$'.

Do deep neural networks evaluate arithmetic expressions using the principle of compositionality in this way?
We investigated this question by analyzing models trained on the Mathematics Dataset \cite{SaxtonGrefenstetteHillEtAl19b}.
This dataset contains 112 million mathematical word problems separated into different modules, each of which covering a different mathematical domain such as arithmetic, algebra, calculus and probability. 
Models receive as input a sequence of characters (e.g., \texttt{Evaluate (12/3 + 10/2)/3}) and must output the sequence of characters exactly matching the correct answer (e.g., \texttt{3}). 
To an untrained model, these strings of characters have no structure or semantic content whatsoever --- nothing in the characters themselves conveys their semantics (e.g., that `2' is larger than `1') or the rules governing them (e.g., the correct order of operations). 

Previous work showed that transformers can achieve an impressive 77.42\% accuracy on this dataset, and that when the standard transformer is augmented with explicitly-structured, tensor product representations \cite{Smolensky90a} --- the TP-Transformer --- this improves to 80.67\% \cite{SchlagSmolenskyFernandezEtAl19b}. 
These models can produce the correct answers to problems that would be challenging even to humans --- for example, problems involving simultaneous differentiation and factorization:

\begin{displayquote}
\textbf{Question}: \texttt{Let r(g) be the second derivative of 2*g**3/3 - 21*g**2/2 + 10*g. Let z be r(7). Factor -z*s + 6 - 9*s**2 + 0*s + 6*s**2}

\textbf{Answer}: \texttt{-(s + 3)*(3*s - 2)}.
\end{displayquote}

However, these models were found to exhibit poor generalization performance on problems with significant deviations from their training set (i.e., ``extrapolation'' problems, including problems with larger numbers, more terms, etc.). This may suggest that although the models were capable of answering unseen word problems as complicated as the one above, they fail to fully capture the systematicity of the rules governing mathematical expressions, possibly relying instead on a ``mix-and-match'' strategy \cite{LakeBaroni18}.

In this work, we analyzed the representations of these trained models, probing them on new problems to investigate whether they evaluated arithmetic expressions compositionally, i.e., whether representations of the partial results corresponding to each sub-expression in the overall question could be found in different parts of the network.
Our results suggest that to a surprising extent, both the standard transformer and the TP-Transformer learn to solve arithmetic problems by evaluating sub-expressions separately, thus demonstrating some ability to compose the meanings of symbols according to their structured relationships.

\section{Methods}

The code used in our analyses can be found online\footnote{\url{https://github.com/jlrussin/interpret-math-transformer}}.

\subsection{Models}

The models we used in our analyses were already trained on the Mathematics Dataset, and were freely available online. 
Details about the architectures and procedures used to train them can be found in the original publication \cite{SchlagSmolenskyFernandezEtAl19b}.
Briefly, both the standard transformer and TP-Transformer contain an encoder that processes the question, and a decoder that generates the final answer. 
The encoder and decoder of both architectures had 6 transformer layers containing multi-head attention modules with 8 heads, as described in \citeA{VaswaniShazeerParmarEtAl17b}. 
Each head in each layer generates a query ($Q$), key ($K$), and value ($V$) vector for every input to that layer. 
Attention distributions are generated by taking a softmax of the scaled dot product of the queries and keys.
The final output of the attention mechanism is the average of the value vectors, weighted by the attention distribution:
\begin{equation}
    \text{Attention}(Q, K, V) = \text{softmax} \left( \frac{Q K^T}{\sqrt{d_k}} \right) V 
\end{equation}
where $d_k$ is the dimension of the key vectors.

The TP-Transformer adapts the transformer architecture to use a role-filler binding mechanism, where roles are meant to explicitly capture structural or relational information in the inputs \cite{SchlagSmolenskyFernandezEtAl19b, Smolensky90a}.
The architecture shares much of the organization of the standard transformer, but an additional role vector ($R$) is generated in each head, and this vector is bound to the existing output of the attention mechanism with a Hadamard product:
\begin{equation}
    \text{TP-Attention}(Q, K, V, R) = \text{softmax} \left( \frac{Q K^T}{\sqrt{d_k}} \right) V \odot R
\end{equation}
where $\odot$ denotes the Hadamard product. 
We included the TP-Transformer in our exploration of compositionality because it had been shown to achieve state-of-the-art performance on the Mathematics Dataset \cite{SchlagSmolenskyFernandezEtAl19b}, and because it was hypothesized to better capture structured relationships. 

Both models were trained on all modules in the Mathematics Dataset \cite{SaxtonGrefenstetteHillEtAl19b} simultaneously and were not fine-tuned on the arithmetic module.
The TP-Transformer used in our analyses had about 49 million parameters and was trained for 1.7 million steps, and the standard transformer had about 44 million parameters and was trained for 700,000 steps. 
These differences were perhaps reflected in the results of the arithmetic module (``arithmetic mixed'') of the dataset, where the TP-Transformer and standard transformer achieved about 83\% and 61\% accuracy, respectively, on the test set. 

\subsection{Test Datasets}

To investigate the degree to which these models had learned to break arithmetic expressions into sub-components, we tested them on problems containing well-defined sub-expressions so that we could systematically probe their internal representations. 
These test sets were derived from the arithmetic module of the dataset, but were highly controlled in a number of ways. 
We created a total of six test sets, and each contained problems that were generated from one of the following arithmetic expressions: 1) $\frac{x_1 + x_2}{x_3}$, 2) $\frac{x_1 * x_2}{x_3 * x_4}$, 3) $x_1 + x_2*x_3$, 4) $\frac{x_1 + x_2*x_3}{x_4}$, 5) $\frac{x_1 + x_2}{x_3} + x_4$, 6) $\frac{x_1 + x_2}{x_3} + \frac{x_4 + x_5}{x_6}$.
These expressions were chosen in order to have a good mix of the possible operations, while retaining unambiguous sub-expressions that we could probe. 
Each test set consisted of roughly 200 problems that were generated by randomly sampling single-digit numbers for each of the $x_i$, while constraining the final answers to be positive integers between 0 and 20. 
This was done so that each number was restricted to a single character, and in order to avoid complications that may have occurred due to negative numbers and arithmetic carrying. 
The samples were also selected such that the distributions of final answers in each test set were as uniform as possible.
The dataset used to train the models did not contain any of the exact samples used for testing, but it did contain some problems with the same arithmetic forms. 
The models that had been trained on the entire dataset were tested on these custom test sets without fine-tuning on them. 

\begin{figure*}[h!]
\centering
    \includegraphics[width=\linewidth]{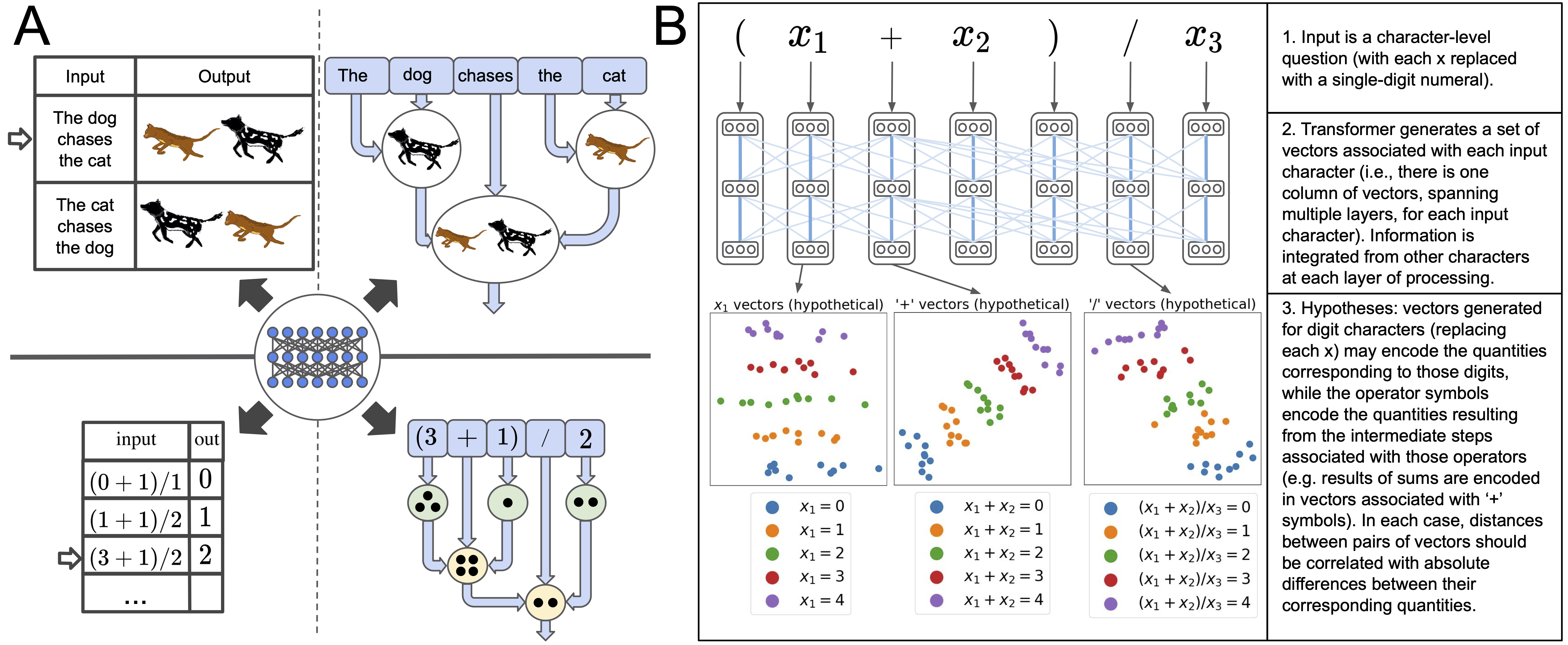}
    \caption{Conceptual illustrations of hypotheses. 
    \textbf{(A)} Two extreme strategies that might be learned by a neural network trained on language tasks (top) or arithmetic problems (bottom). 
    A large model might memorize in its weights a lookup table containing answers to each problem (i.e., a holistic pattern-matching strategy), thus bypassing their implicit compositionality altogether (left). 
    Alternatively, the model may infer and leverage the underlying structure of the problem to decompose it into sub-expressions (right). 
    In the latter case, the model's representations should show evidence of encoding the meaning of each symbol (e.g., quantities corresponding to numerals, shown in green circles) and the meaning of each sub-expression (e.g., quantities corresponding to results of intermediate operations, shown in yellow circles).
    \textbf{(B)} Hypothetical compositional representations in a transformer model trained on math problems. The vectors representing digit characters encode the values of the corresponding digits, in the sense that vectors associated with similar values are closer together. Similarly, the vectors representing operators encode the intermediate results of those operations. Distances between pairs of vectors should be highly correlated with the absolute differences between their corresponding values. Only three layers are shown, but both models had six layers.}
    \label{fig:fig1}
\end{figure*}

The models generally achieved high accuracies on the probing test sets we created (numbers corresponding to expressions above): TP-Transformer 1) 100\%, 2) 100\%, 3) 92\%, 4) 89.5\%, 5) 100\%, 6) 96\%; standard transformer: 1) 100\%, 2) 100\%, 3) 38\%, 4) 98.5\%, 5) 100\%, 6) 98\%. 
Differences in accuracy on the test sets were not critical to our analyses, as our results were qualitatively reproduced for both models.
The problems incorrectly answered by the models were not excluded from the analyses.

\section{Results}

Two extreme strategies for solving problems with implicit compositional structure are shown in Figure \ref{fig:fig1}A. At one extreme, a large neural network might overlook the compositional structure of the problems and memorize in its weights a simple lookup table containing the answers to each problem individually. At the other extreme, a perfectly compositional learner would infer whatever structured relationships exist and use them to decompose problems into appropriate sub-components. 

We hypothesized that if the models had learned to evaluate arithmetic expressions by processing their sub-components separately, then the representations corresponding to similar quantities would be closer together, as measured by Euclidean distance (see Figure \ref{fig:fig1}B). 
For example, in expressions of the form $\frac{x_1 + x_2}{x_3}$ (where each $x_i$ is replaced by a particular numeral in each problem), the vectors corresponding to the partial result in the numerator $x_1 + x_2 = 1$ would be closer to those corresponding to $x_1 + x_2 = 2$ than they would be to those corresponding to $x_1 + x_2 = 3$.
We therefore extracted multiple vectors (e.g., queries, keys, values, and role vectors for TP-Transformer) from each layer of both models in order to analyze them.
Our analyses across these different kinds of vectors from each attention head yielded qualitatively similar results, so for simplicity we report results for queries across both models. 
Unless noted otherwise, we report results from vectors extracted from the highest layer (layer 6) of the encoder of each model.

\begin{savenotes}
\begin{table}[h!] 
\centering
\caption{Spearman correlations for operators in each test set. All were highly significant ($p < 0.001$). Vectors were from the last layer of the encoder.}
\begin{tabular}{llcc}
\toprule
Expression & Operator & TP & TF \\
\midrule
1. $(x_1 + x_2)/x_3$                    & `$+$'       & .315 & .453 \\
                                        & `$/$'       & .565 & .539 \\
\midrule 
2. $(x_1 * x_2)/(x_3 * x_4)$            & `$*$' (1st) & .479 & .612 \\
                                        & `$*$' (2nd) & .662 & .654 \\
                                        & `$/$'       & .590 & .536 \\
\midrule
3. $x_1 + x_2 * x_3$                    & `$*$'       & .132 & .251 \\
                                        & `$+$'       & .502 & .381 \\
\midrule
4. $(x_1 + x_2*x_3)/x_4$                & `$*$'       & .141 & .297 \\
                                        & `$+$'       & .159 & .174 \\
                                        & `$/$'       & .147 & .171 \\
\midrule
5. $(x_1 + x_2)/x_3 + x_4 $             & `$+$' (1st) & .424 & .423 \\
                                        & `$/$'       & .510 & .459 \\
                                        & `$+$' (2nd) & .353 & .411 \\
\midrule
6. $(x_1 + x_2)/x_3 + (x_4 + x_5)/x_6$  & `$+$' (1st) & .410 & .393 \\
                                        & `$/$' (1st) & .610 & .536 \\
                                        & `$+$' (2nd) & .405 & .455 \\
                                        & `$/$' (2nd) & .566 & .526 \\
                                        & `$+$' (3rd) & .303 & .421 \\
\bottomrule
\end{tabular}
\label{tab:tab1}
\end{table}
\end{savenotes}

\begin{figure*}[h!]
\centering
    \includegraphics[width=\linewidth]{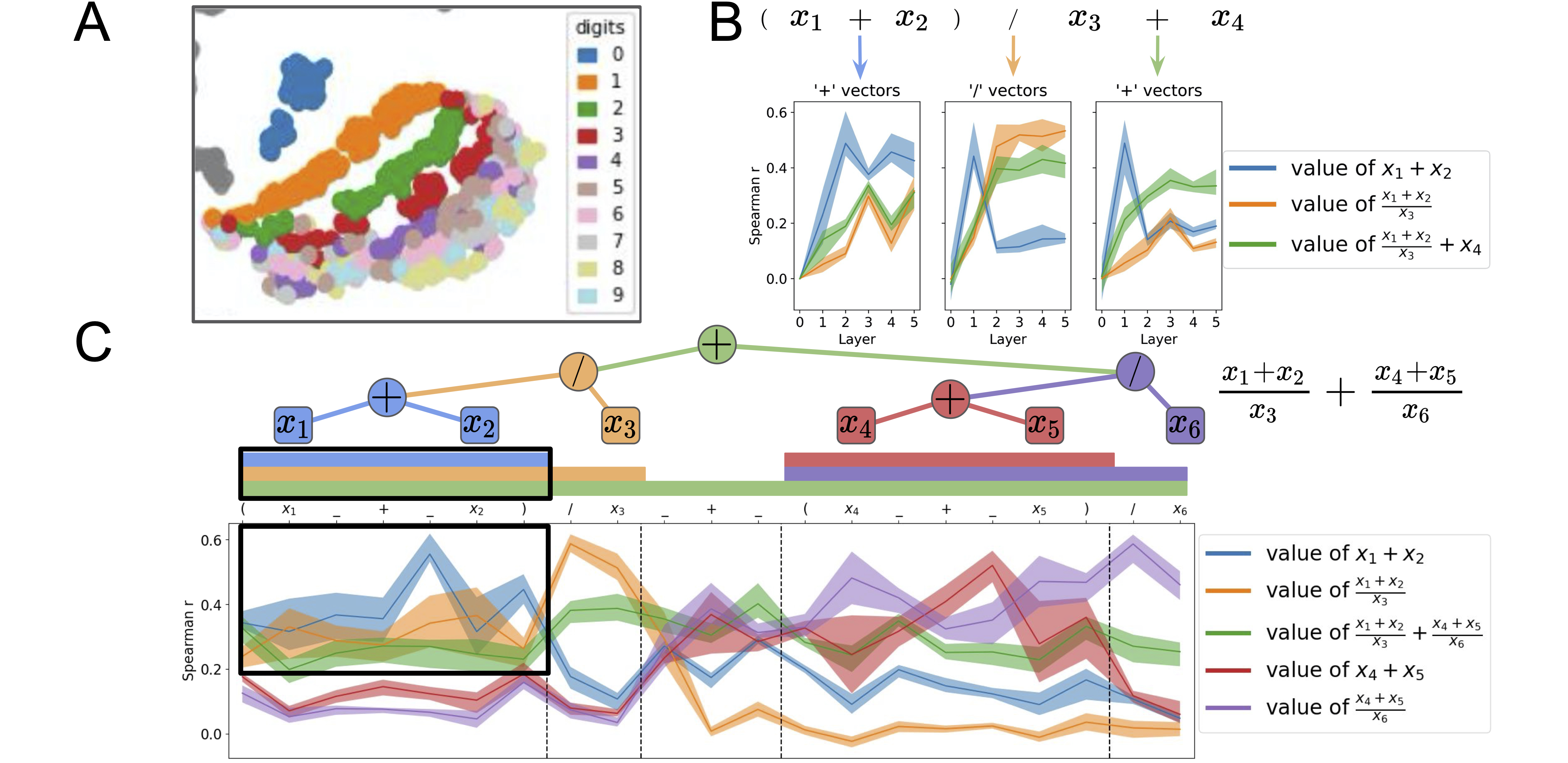}
    \caption{Results. \textbf{(A)} t-SNE embedding of vectors representing digits in layer 1 of the TP-Transformer on the test set of the mixed arithmetic module. Each point designates a vector associated with a particular digit character, colored by its value. The semantics of the digit characters is apparent in the model's representations, which are partially organized according to their natural order (along the top-left to bottom-right axis). \textbf{(B)} Correlations with vectors associated with operators in TP-Transformer. In each problem, each of the $x_i$ were replaced with single-digit numerals. Vectors representing the first `$+$' symbol encode the results of the sum in the numerator ($x_1 + x_2$), vectors representing the `$/$' symbol encode the results of the division ($(x_1 + x_2)/x_3$), and vectors representing the second `$+$' symbol encode the results of the final sum ($(x_1 + x_2)/x_3 + x_4$). In general, analyses focused on the vectors generated in the final layer, but correlations displayed here across all layers show how these can change with further processing (e.g., representations of the `$/$' symbol seem to encode the result of the sum in the first layer, but in later layers strongly encode the result of the division). Colored envelopes show the minimum and maximum correlation across heads. \textbf{(C)} Alignment of correlation measures across the entire input sequence with the parse tree of the expression, colored to match the lines in the plot for the appropriate quantities. Correlations with the values of an intermediate result peak at the vectors associated with the corresponding operator, and are elevated over the constituent containing its arguments (shown by matching colors in the parse tree, and colored bars covering the extent of the appropriate constituent). Black boxes highlight an example: the vectors representing each element in the constituent $(x_1 + x_2)$ show the highest correlations with the value of the sum of $x_1$ and $x_2$ (blue), but also show correlations with its division by $x_3$ (orange), and the result of the whole expression (green). Results are from vectors in the final layer of the TP-Transformer encoder. Vertical dotted lines delineate sub-expressions, and underscores indicate space characters.}
    \label{fig:fig2}
\end{figure*}

As a first step, we probed the models for evidence that they were encoding the quantities associated with digit characters (akin to the dots shown on the right of Figure \ref{fig:fig1}A).
Figure \ref{fig:fig2}A shows the vectors representing digits in layer 1 of the TP-Transformer model, visualized using t-SNE \cite{vanderMaatenHinton08}. 
The model's representations capture the semantics of the digit characters in that they are organized according to their natural order (e.g., vectors corresponding to 1's are closer to 2's than to 9's, etc.).
We confirmed this pattern quantitatively by measuring Spearman correlations of the distances between pairs of digit vectors with the absolute differences between the values of their corresponding digits.
This correlation was significant for both TP-Transformer (TP: $r = 0.440, p < 0.001$), and the standard transformer (TF: $r = 0.436, p < 0.001$). 
These were compared to correlations between these same distances and the absolute differences between the values of the \emph{other} digits in the same problems. 
For example, in problems of the form $\frac{x_1 + x_2}{x_3}$, the distances in this ``unmatched'' correlation would be taken between vectors corresponding to $x_1$ vectors, but the absolute differences would be taken between the values of $x_3$ from the same problems. 
This further analysis revealed that the matched correlations (TP: $r = 0.440$, TF: $r = 0.436$) were much higher than the unmatched (TP: $r = 0.086$, TF: $r = 0.087$), indicating that the models were representing the quantities associated with each numeral in the appropriate position in the sequence. 
This relationship was found to be statistically significant in a more formal linear regression comparing ``matched'' and ``unmatched'' pairs ($p < 0.001$ for both models). 

Next, we investigated whether the encoding of intermediate results of arithmetic sub-expressions could be detected in a similar way.
We reasoned that if this pattern of correlation were observed for the intermediate results corresponding to arithmetic sub-expressions, this would indicate that the model had encoded these partial results. 
We repeated the analyses, but with distances between vectors associated with operators and the differences between their corresponding intermediate results. 
For example, if the vectors representing the `$+$' symbol in expressions with the form $\frac{x_1 + x_2}{x_3}$ were encoding the sum of $x_1$ and $x_2$, we would expect those `$+$' vectors encoding similar values for their sums to be closer together, leading to a significant correlation between distances between the pairs of `$+$' vectors and the differences between their corresponding sums. 
This correlation would be expected to peak at the position of the operator, but also be elevated over the positions of symbols within its constituent (e.g., in $(x_1 + x_2)/x_3$, the positions over the $x_1 + x_2$ constituent for the `+' operator).

These analyses revealed a striking pattern: transformer models trained on mathematics encode the quantities of intermediate results in the vectors associated with the appropriate operators (see Table \ref{tab:tab1}). 
Figure \ref{fig:fig2}B shows how these correlations unfold over the layers of the network, with comparisons to the inappropriate operators (e.g., `$+$' vectors and the result of the division).
When these data were aggregated across all the expressions we tested, a significant correlation was observed between corresponding operator representations and partial values for both models (TP: $r = 0.310, p < 0.001$; TF: $r = 0.314, p < 0.001$).
This correlation was higher than when distances were correlated with differences between intermediate results corresponding to the other operators in the same problems (TP: $r = 0.122$, TF: $r = 0.167$).
Again, a more formal linear regression revealed that this relationship was statistically significant ($p < 0.001$ for both models), indicating that the models were representing these intermediate results in the vectors corresponding to the appropriate operators. 
Figure \ref{fig:fig2}C shows the same correlation measures on representations across the entire input sequence of a more complicated expression.
The correlations with partial results are seen to peak at the corresponding operators, but also to be relatively elevated over the corresponding constituents.
We repeated our analyses and confirmed that the vectors representing symbols within constituents also carried information about their corresponding partial results (TP: $r = 0.201$, TF: $r = 0.189$). 

\section{Discussion}

Compositionality, a hallmark of human cognition, requires knowledge of constituent structure to guide the assembly of partial meanings into coherent semantic wholes.
Our analyses reveal that the compositional semantics implicit in character-level math problems can emerge to a surprising extent in neural networks even when they are instructed only on the characters comprising the problems' answers.
Our results are consistent with previous work in the natural language setting \cite{ManningClarkHewittEtAl20a, TenneyDasPavlick19a} suggesting that when these networks are trained on many observations, they can learn to represent structured relationships.
However, our work shows further that these models can use this knowledge to guide a partially compositional process whereby semantic content is integrated across symbols.

Though our results were surprising, we expect that the compositionality we observed is imperfect, and that the trained models lie somewhere between the two extremes depicted in Figure \ref{fig:fig1}A.
A perfectly compositional learner would completely separate the process of evaluating each of the sub-components of an arithmetic expression (e.g., first completely evaluate the expression in the numerator, then divide the result by the denominator). 
Our results suggest that transformers perform these separable computations in different parts of the network: the vectors aligned with the input positions of the operators tended to encode the quantities corresponding to the results of those operations.
However, our analyses did not find that these relationships were perfect (e.g., see the non-zero correlations across the sequence in Figure \ref{fig:fig2}C). 
Furthermore, it is likely that these vectors are encoding more than pure quantities; the high-dimensional vectors in these models may encode the sum ($x_1 + x_2$) while also encoding each of the constituent elements (such that a decoder could be trained to predict the constituents, as well as the sum, from the vector).

Previous work on transformers trained on the Mathematics Dataset \cite{SchlagSmolenskyFernandezEtAl19b} showed that these models suffered large reductions in generalization performance on arithmetic problems with significantly different surface-level features (e.g., problems with larger numbers, more terms, etc.). 
A perfectly compositional agent with a true understanding of arithmetic rules would in principle be able to generalize to any problem following those rules.
It is possible that although our analyses revealed a significant degree of compositionality, its imperfection prevented the models from generalizing on these extrapolation tests.

It should be noted that although compositionality allows for flexible cognition and a powerful form of combinatorial generalization \cite{fodor1988connectionism, LakeUllmanTenenbaumEtAl17}, a strict form of compositionality may not always be desirable --- for example, when learning the meanings of idioms or other idiosyncrasies of natural language that violate the principle of compositionality \cite{Szabo20}.
Human language learners must negotiate the tension between a strict compositionality assumption and the ability to learn exceptions \cite{RumelhartMcClelland86a, PinkerPrince88}.
Machine learning methods for natural language processing face a similar tension, and may benefit from a greater dialogue with ongoing research in cognitive science \cite{LakeUllmanTenenbaumEtAl17}.

The success of modern neural networks has shown that major advances in artificial intelligence are possible when large models are trained on enough data \cite{BrownMannRyderEtAl20, LeCunBengioHinton15}.
Our results show that compositionality, which is often thought to be an inherent property of human cognition \cite{fodor1988connectionism}, can to some extent emerge when a large neural network is trained on enough data.
However, much work remains to clarify whether these approaches will continue to scale to human-level compositionality, or whether this will require learning systems that have been explicitly designed to facilitate compositional processing \cite{Smolensky90a, RussinOReillyBengio20a}.

\subsection{Acknowledgments}
We would like to thank Imanol Schlag, J\"urgen Schmidhuber, Randall O'Reilly, the members of the Computational Cognitive Neuroscience lab at UC Davis, and reviewers for helpful comments and discussions. J.R. was supported by the NIMH, Award Number T32MH112507. The content is solely the responsibility of the authors and does not necessarily represent the official views of the NIH. Illustrations by H. Tomkiewicz.

\bibliographystyle{apacite}

\setlength{\bibleftmargin}{.125in}
\setlength{\bibindent}{-\bibleftmargin}

\bibliography{cogsci2021}

\begin{thebibliography}{}

\bibitem [\protect \citeauthoryear {%
Brown%
\ \protect \BOthers {.}}{%
Brown%
\ \protect \BOthers {.}}{%
{\protect \APACyear {2020}}%
}]{%
BrownMannRyderEtAl20}
\APACinsertmetastar {%
BrownMannRyderEtAl20}%
\begin{APACrefauthors}%
Brown, T\BPBI B.%
, Mann, B.%
, Ryder, N.%
, Subbiah, M.%
, Kaplan, J.%
, Dhariwal, P.%
\BDBL {}Amodei, D.%
\end{APACrefauthors}%
\unskip\
\newblock
\APACrefYearMonthDay{2020}{{\APACmonth{05}}}{}.
\newblock
{\BBOQ}\APACrefatitle {Language {{Models}} Are {{Few}}-{{Shot Learners}}}
  {Language {{Models}} are {{Few}}-{{Shot Learners}}}.{\BBCQ}
\newblock

\PrintBackRefs{\CurrentBib}

\bibitem [\protect \citeauthoryear {%
Devlin%
, Chang%
, Lee%
\BCBL {}\ \BBA {} Toutanova%
}{%
Devlin%
\ \protect \BOthers {.}}{%
{\protect \APACyear {2019}}%
}]{%
DevlinChangLeeEtAl19b}
\APACinsertmetastar {%
DevlinChangLeeEtAl19b}%
\begin{APACrefauthors}%
Devlin, J.%
, Chang, M\BHBI W.%
, Lee, K.%
\BCBL {}\ \BBA {} Toutanova, K.%
\end{APACrefauthors}%
\unskip\
\newblock
\APACrefYearMonthDay{2019}{}{}.
\newblock
{\BBOQ}\APACrefatitle {{{BERT}}: {{Pre}}-Training of Deep Bidirectional
  Transformers for Language Understanding} {{{BERT}}: {{Pre}}-training of deep
  bidirectional transformers for language understanding}.{\BBCQ}
\newblock
\BIn{} J.~Burstein, C.~Doran\BCBL {}\ \BBA {} T.~Solorio\ (\BEDS),
  \APACrefbtitle {Proc. Conf. {{NA}} Chapt. Assoc. for Comp. Ling.} {Proc.
  conf. {{NA}} chapt. assoc. for comp. ling.}\ (\BPGS\ 4171--4186).
\newblock
\APACaddressPublisher{{Minneapolis, MN, USA}}{{ACL}}.
\newblock
\begin{APACrefDOI} \doi{10.18653/v1/n19-1423} \end{APACrefDOI}
\PrintBackRefs{\CurrentBib}

\bibitem [\protect \citeauthoryear {%
Fodor%
\ \BBA {} Pylyshyn%
}{%
Fodor%
\ \BBA {} Pylyshyn%
}{%
{\protect \APACyear {1988}}%
}]{%
fodor1988connectionism}
\APACinsertmetastar {%
fodor1988connectionism}%
\begin{APACrefauthors}%
Fodor, J\BPBI A.%
\BCBT {}\ \BBA {} Pylyshyn, Z\BPBI W.%
\end{APACrefauthors}%
\unskip\
\newblock
\APACrefYearMonthDay{1988}{}{}.
\newblock
{\BBOQ}\APACrefatitle {Connectionism and cognitive architecture: A critical
  analysis} {Connectionism and cognitive architecture: A critical
  analysis}.{\BBCQ}
\newblock
\APACjournalVolNumPages{Cognition}{28}{1-2}{3--71}.
\PrintBackRefs{\CurrentBib}

\bibitem [\protect \citeauthoryear {%
Hinzen%
, Machery%
\BCBL {}\ \BBA {} Werning%
}{%
Hinzen%
\ \protect \BOthers {.}}{%
{\protect \APACyear {2012}}%
}]{%
hinzen2012oxford}
\APACinsertmetastar {%
hinzen2012oxford}%
\begin{APACrefauthors}%
Hinzen, W.%
, Machery, E.%
\BCBL {}\ \BBA {} Werning, M.%
\end{APACrefauthors}%
\ (\BEDS).
\unskip\
\newblock
\APACrefYear{2012}.
\newblock
\APACrefbtitle {The Oxford Handbook of Compositionality} {The oxford handbook
  of compositionality}.
\newblock
\APACaddressPublisher{}{Oxford University Press}.
\PrintBackRefs{\CurrentBib}

\bibitem [\protect \citeauthoryear {%
Lake%
\ \BBA {} Baroni%
}{%
Lake%
\ \BBA {} Baroni%
}{%
{\protect \APACyear {2018}}%
}]{%
LakeBaroni18}
\APACinsertmetastar {%
LakeBaroni18}%
\begin{APACrefauthors}%
Lake, B\BPBI M.%
\BCBT {}\ \BBA {} Baroni, M.%
\end{APACrefauthors}%
\unskip\
\newblock
\APACrefYearMonthDay{2018}{}{}.
\newblock
{\BBOQ}\APACrefatitle {Generalization without Systematicity: {{On}} the
  Compositional Skills of Sequence-to-Sequence Recurrent Networks}
  {Generalization without systematicity: {{On}} the compositional skills of
  sequence-to-sequence recurrent networks}.{\BBCQ}
\newblock
\BIn{} J\BPBI G.~Dy\ \BBA {} A.~Krause\ (\BEDS), \APACrefbtitle {Proc. of the
  35th {{Intern}}. {{Conf}}. on {{Mach}}. {{Lear}}.} {Proc. of the 35th
  {{Intern}}. {{Conf}}. on {{Mach}}. {{Lear}}.}\ (\BVOL~80, \BPGS\ 2879--2888).
\newblock
\APACaddressPublisher{{Stockholmsm\"assan, Stockholm, Sweden}}{{PMLR}}.
\PrintBackRefs{\CurrentBib}

\bibitem [\protect \citeauthoryear {%
Lake%
, Ullman%
, Tenenbaum%
\BCBL {}\ \BBA {} Gershman%
}{%
Lake%
\ \protect \BOthers {.}}{%
{\protect \APACyear {2017}}%
}]{%
LakeUllmanTenenbaumEtAl17}
\APACinsertmetastar {%
LakeUllmanTenenbaumEtAl17}%
\begin{APACrefauthors}%
Lake, B\BPBI M.%
, Ullman, T\BPBI D.%
, Tenenbaum, J\BPBI B.%
\BCBL {}\ \BBA {} Gershman, S\BPBI J.%
\end{APACrefauthors}%
\unskip\
\newblock
\APACrefYearMonthDay{2017}{{\APACmonth{01}}}{}.
\newblock
{\BBOQ}\APACrefatitle {Building Machines That Learn and Think like People}
  {Building machines that learn and think like people}.{\BBCQ}
\newblock
\APACjournalVolNumPages{The Behavioral and Brain Sciences}{40}{}{e253}.
\newblock
\begin{APACrefDOI} \doi{10.1017/S0140525X16001837} \end{APACrefDOI}
\PrintBackRefs{\CurrentBib}

\bibitem [\protect \citeauthoryear {%
LeCun%
, Bengio%
\BCBL {}\ \BBA {} Hinton%
}{%
LeCun%
\ \protect \BOthers {.}}{%
{\protect \APACyear {2015}}%
}]{%
LeCunBengioHinton15}
\APACinsertmetastar {%
LeCunBengioHinton15}%
\begin{APACrefauthors}%
LeCun, Y.%
, Bengio, Y.%
\BCBL {}\ \BBA {} Hinton, G.%
\end{APACrefauthors}%
\unskip\
\newblock
\APACrefYearMonthDay{2015}{{\APACmonth{05}}}{}.
\newblock
{\BBOQ}\APACrefatitle {Deep Learning} {Deep learning}.{\BBCQ}
\newblock
\APACjournalVolNumPages{Nature}{521}{7553}{436--444}.
\newblock
\begin{APACrefDOI} \doi{10.1038/nature14539} \end{APACrefDOI}
\PrintBackRefs{\CurrentBib}

\bibitem [\protect \citeauthoryear {%
Linzen%
\ \BBA {} Baroni%
}{%
Linzen%
\ \BBA {} Baroni%
}{%
{\protect \APACyear {2021}}%
}]{%
LinzenBaroni21}
\APACinsertmetastar {%
LinzenBaroni21}%
\begin{APACrefauthors}%
Linzen, T.%
\BCBT {}\ \BBA {} Baroni, M.%
\end{APACrefauthors}%
\unskip\
\newblock
\APACrefYearMonthDay{2021}{}{}.
\newblock
{\BBOQ}\APACrefatitle {Syntactic {{Structure}} from {{Deep Learning}}}
  {Syntactic {{Structure}} from {{Deep Learning}}}.{\BBCQ}
\newblock
\APACjournalVolNumPages{Annual Review of Linguistics}{7}{1}{}.
\newblock
\begin{APACrefDOI} \doi{10.1146/annurev-linguistics-032020-051035}
  \end{APACrefDOI}
\PrintBackRefs{\CurrentBib}

\bibitem [\protect \citeauthoryear {%
Manning%
, Clark%
, Hewitt%
, Khandelwal%
\BCBL {}\ \BBA {} Levy%
}{%
Manning%
\ \protect \BOthers {.}}{%
{\protect \APACyear {2020}}%
}]{%
ManningClarkHewittEtAl20a}
\APACinsertmetastar {%
ManningClarkHewittEtAl20a}%
\begin{APACrefauthors}%
Manning, C\BPBI D.%
, Clark, K.%
, Hewitt, J.%
, Khandelwal, U.%
\BCBL {}\ \BBA {} Levy, O.%
\end{APACrefauthors}%
\unskip\
\newblock
\APACrefYearMonthDay{2020}{{\APACmonth{06}}}{}.
\newblock
{\BBOQ}\APACrefatitle {Emergent Linguistic Structure in Artificial Neural
  Networks Trained by Self-Supervision} {Emergent linguistic structure in
  artificial neural networks trained by self-supervision}.{\BBCQ}
\newblock
\APACjournalVolNumPages{Proceedings of the National Academy of
  Sciences}{}{}{201907367}.
\newblock
\begin{APACrefDOI} \doi{10.1073/pnas.1907367117} \end{APACrefDOI}
\PrintBackRefs{\CurrentBib}

\bibitem [\protect \citeauthoryear {%
Martin%
\ \BBA {} Baggio%
}{%
Martin%
\ \BBA {} Baggio%
}{%
{\protect \APACyear {2020}}%
}]{%
MartinBaggio20a}
\APACinsertmetastar {%
MartinBaggio20a}%
\begin{APACrefauthors}%
Martin, A\BPBI E.%
\BCBT {}\ \BBA {} Baggio, G.%
\end{APACrefauthors}%
\unskip\
\newblock
\APACrefYearMonthDay{2020}{{\APACmonth{02}}}{}.
\newblock
{\BBOQ}\APACrefatitle {Modelling Meaning Composition from Formalism to
  Mechanism} {Modelling meaning composition from formalism to
  mechanism}.{\BBCQ}
\newblock
\APACjournalVolNumPages{Philosophical Transactions of the Royal Society B:
  Biological Sciences}{375}{1791}{20190298}.
\newblock
\begin{APACrefDOI} \doi{10.1098/rstb.2019.0298} \end{APACrefDOI}
\PrintBackRefs{\CurrentBib}

\bibitem [\protect \citeauthoryear {%
McClelland%
, Hill%
, Rudolph%
, Baldridge%
\BCBL {}\ \BBA {} Sch{\"u}tze%
}{%
McClelland%
\ \protect \BOthers {.}}{%
{\protect \APACyear {2020}}%
}]{%
McClellandHillRudolphEtAl20a}
\APACinsertmetastar {%
McClellandHillRudolphEtAl20a}%
\begin{APACrefauthors}%
McClelland, J\BPBI L.%
, Hill, F.%
, Rudolph, M.%
, Baldridge, J.%
\BCBL {}\ \BBA {} Sch{\"u}tze, H.%
\end{APACrefauthors}%
\unskip\
\newblock
\APACrefYearMonthDay{2020}{{\APACmonth{10}}}{}.
\newblock
{\BBOQ}\APACrefatitle {Placing Language in an Integrated Understanding System:
  {{Next}} Steps toward Human-Level Performance in Neural Language Models}
  {Placing language in an integrated understanding system: {{Next}} steps
  toward human-level performance in neural language models}.{\BBCQ}
\newblock
\APACjournalVolNumPages{Proceedings of the National Academy of
  Sciences}{117}{42}{25966--25974}.
\newblock
\begin{APACrefDOI} \doi{10.1073/pnas.1910416117} \end{APACrefDOI}
\PrintBackRefs{\CurrentBib}

\bibitem [\protect \citeauthoryear {%
Pinker%
\ \BBA {} Prince%
}{%
Pinker%
\ \BBA {} Prince%
}{%
{\protect \APACyear {1988}}%
}]{%
PinkerPrince88}
\APACinsertmetastar {%
PinkerPrince88}%
\begin{APACrefauthors}%
Pinker, S.%
\BCBT {}\ \BBA {} Prince, A.%
\end{APACrefauthors}%
\unskip\
\newblock
\APACrefYearMonthDay{1988}{{\APACmonth{03}}}{}.
\newblock
{\BBOQ}\APACrefatitle {On Language and Connectionism: {{Analysis}} of a
  Parallel Distributed Processing Model of Language Acquisition} {On language
  and connectionism: {{Analysis}} of a parallel distributed processing model of
  language acquisition}.{\BBCQ}
\newblock
\APACjournalVolNumPages{Cognition}{28}{1}{73--193}.
\newblock
\begin{APACrefDOI} \doi{10.1016/0010-0277(88)90032-7} \end{APACrefDOI}
\PrintBackRefs{\CurrentBib}

\bibitem [\protect \citeauthoryear {%
Rumelhart%
\ \BBA {} McClelland%
}{%
Rumelhart%
\ \BBA {} McClelland%
}{%
{\protect \APACyear {1986}}%
}]{%
RumelhartMcClelland86a}
\APACinsertmetastar {%
RumelhartMcClelland86a}%
\begin{APACrefauthors}%
Rumelhart, D\BPBI E.%
\BCBT {}\ \BBA {} McClelland, J\BPBI L.%
\end{APACrefauthors}%
\unskip\
\newblock
\APACrefYearMonthDay{1986}{{\APACmonth{01}}}{}.
\newblock
{\BBOQ}\APACrefatitle {On Learning the Past Tenses of English Verbs} {On
  learning the past tenses of english verbs}.{\BBCQ}
\newblock
\BIn{} J\BPBI L.~McClelland, D\BPBI E.~Rumelhart\BCBL {}\ \BBA {} P\BPBI
  R.~Group\ (\BEDS), \APACrefbtitle {Parallel {{Distributed Processing}}.
  {{Volume}} 2: {{Psychological}} and {{Biological Models}}} {Parallel
  {{Distributed Processing}}. {{Volume}} 2: {{Psychological}} and {{Biological
  Models}}}\ (\BPGS\ 216--271).
\newblock
\APACaddressPublisher{{Cambridge, MA}}{{MIT Press}}.
\PrintBackRefs{\CurrentBib}

\bibitem [\protect \citeauthoryear {%
Russin%
, O'Reilly%
\BCBL {}\ \BBA {} Bengio%
}{%
Russin%
\ \protect \BOthers {.}}{%
{\protect \APACyear {2020}}%
}]{%
RussinOReillyBengio20a}
\APACinsertmetastar {%
RussinOReillyBengio20a}%
\begin{APACrefauthors}%
Russin, J.%
, O'Reilly, R\BPBI C.%
\BCBL {}\ \BBA {} Bengio, Y.%
\end{APACrefauthors}%
\unskip\
\newblock
\APACrefYearMonthDay{2020}{}{}.
\newblock
{\BBOQ}\APACrefatitle {Deep Learning Needs a Prefrontal Cortex} {Deep learning
  needs a prefrontal cortex}.{\BBCQ}
\newblock
\BIn{} \APACrefbtitle {Bridging {{AI}} and {{Cognitive Science}} ({{BAICS}})
  {{Workshop}}, {{ICLR}} 2020} {Bridging {{AI}} and {{Cognitive Science}}
  ({{BAICS}}) {{Workshop}}, {{ICLR}} 2020}\ (\BPG~11).
\PrintBackRefs{\CurrentBib}

\bibitem [\protect \citeauthoryear {%
Saxton%
, Grefenstette%
, Hill%
\BCBL {}\ \BBA {} Kohli%
}{%
Saxton%
\ \protect \BOthers {.}}{%
{\protect \APACyear {2019}}%
}]{%
SaxtonGrefenstetteHillEtAl19b}
\APACinsertmetastar {%
SaxtonGrefenstetteHillEtAl19b}%
\begin{APACrefauthors}%
Saxton, D.%
, Grefenstette, E.%
, Hill, F.%
\BCBL {}\ \BBA {} Kohli, P.%
\end{APACrefauthors}%
\unskip\
\newblock
\APACrefYearMonthDay{2019}{}{}.
\newblock
{\BBOQ}\APACrefatitle {Analysing Mathematical Reasoning Abilities of Neural
  Models} {Analysing mathematical reasoning abilities of neural models}.{\BBCQ}
\newblock
\BIn{} \APACrefbtitle {7th {{Intern}}. {{Conf}}. on {{Lear}}. {{Repr}}.} {7th
  {{Intern}}. {{Conf}}. on {{Lear}}. {{Repr}}.}
\newblock
\APACaddressPublisher{{new orleans, LA, USA}}{{OpenReview.net}}.
\PrintBackRefs{\CurrentBib}

\bibitem [\protect \citeauthoryear {%
Schlag%
\ \protect \BOthers {.}}{%
Schlag%
\ \protect \BOthers {.}}{%
{\protect \APACyear {2019}}%
}]{%
SchlagSmolenskyFernandezEtAl19b}
\APACinsertmetastar {%
SchlagSmolenskyFernandezEtAl19b}%
\begin{APACrefauthors}%
Schlag, I.%
, Smolensky, P.%
, Fernandez, R.%
, Jojic, N.%
, Schmidhuber, J.%
\BCBL {}\ \BBA {} Gao, J.%
\end{APACrefauthors}%
\unskip\
\newblock
\APACrefYearMonthDay{2019}{}{}.
\newblock
{\BBOQ}\APACrefatitle {Enhancing the Transformer with Explicit Relational
  Encoding for Math Problem Solving} {Enhancing the transformer with explicit
  relational encoding for math problem solving}.{\BBCQ}
\newblock
\APACjournalVolNumPages{CoRR}{abs/1910.06611}{}{}.
\PrintBackRefs{\CurrentBib}

\bibitem [\protect \citeauthoryear {%
Smolensky%
}{%
Smolensky%
}{%
{\protect \APACyear {1990}}%
}]{%
Smolensky90a}
\APACinsertmetastar {%
Smolensky90a}%
\begin{APACrefauthors}%
Smolensky, P.%
\end{APACrefauthors}%
\unskip\
\newblock
\APACrefYearMonthDay{1990}{{\APACmonth{11}}}{}.
\newblock
{\BBOQ}\APACrefatitle {Tensor Product Variable Binding and the Representation
  of Symbolic Structures in Connectionist Systems} {Tensor product variable
  binding and the representation of symbolic structures in connectionist
  systems}.{\BBCQ}
\newblock
\APACjournalVolNumPages{Artificial Intelligence}{46}{1-2}{159--216}.
\newblock
\begin{APACrefDOI} \doi{10.1016/0004-3702(90)90007-M} \end{APACrefDOI}
\PrintBackRefs{\CurrentBib}

\bibitem [\protect \citeauthoryear {%
Szab{\'o}%
}{%
Szab{\'o}%
}{%
{\protect \APACyear {2020}}%
}]{%
Szabo20}
\APACinsertmetastar {%
Szabo20}%
\begin{APACrefauthors}%
Szab{\'o}, Z\BPBI G.%
\end{APACrefauthors}%
\unskip\
\newblock
\APACrefYearMonthDay{2020}{}{}.
\newblock
{\BBOQ}\APACrefatitle {Compositionality} {Compositionality}.{\BBCQ}
\newblock
\BIn{} E\BPBI N.~Zalta\ (\BED), \APACrefbtitle {The {{Stanford Encyclopedia}}
  of {{Philosophy}}} {The {{Stanford Encyclopedia}} of {{Philosophy}}}\
  (\PrintOrdinal{Fall 2020}\ \BEd).
\newblock
\APACaddressPublisher{}{{Metaphysics Research Lab, Stanford University}}.
\PrintBackRefs{\CurrentBib}

\bibitem [\protect \citeauthoryear {%
Tenney%
, Das%
\BCBL {}\ \BBA {} Pavlick%
}{%
Tenney%
\ \protect \BOthers {.}}{%
{\protect \APACyear {2019}}%
}]{%
TenneyDasPavlick19a}
\APACinsertmetastar {%
TenneyDasPavlick19a}%
\begin{APACrefauthors}%
Tenney, I.%
, Das, D.%
\BCBL {}\ \BBA {} Pavlick, E.%
\end{APACrefauthors}%
\unskip\
\newblock
\APACrefYearMonthDay{2019}{}{}.
\newblock
{\BBOQ}\APACrefatitle {{{BERT}} Rediscovers the Classical {{NLP}} Pipeline}
  {{{BERT}} rediscovers the classical {{NLP}} pipeline}.{\BBCQ}
\newblock
\BIn{} A.~Korhonen, D\BPBI R.~Traum\BCBL {}\ \BBA {} L.~M{\`a}rquez\ (\BEDS),
  \APACrefbtitle {Proc. of the 57th Conf. of the Asso. for Comp. Ling.} {Proc.
  of the 57th conf. of the asso. for comp. ling.}\ (\BPGS\ 4593--4601).
\newblock
\APACaddressPublisher{{Florence, Italy}}{{Association for Computational
  Linguistics}}.
\newblock
\begin{APACrefDOI} \doi{10.18653/v1/p19-1452} \end{APACrefDOI}
\PrintBackRefs{\CurrentBib}

\bibitem [\protect \citeauthoryear {%
{van der Maaten}%
\ \BBA {} Hinton%
}{%
{van der Maaten}%
\ \BBA {} Hinton%
}{%
{\protect \APACyear {2008}}%
}]{%
vanderMaatenHinton08}
\APACinsertmetastar {%
vanderMaatenHinton08}%
\begin{APACrefauthors}%
{van der Maaten}, L.%
\BCBT {}\ \BBA {} Hinton, G.%
\end{APACrefauthors}%
\unskip\
\newblock
\APACrefYearMonthDay{2008}{}{}.
\newblock
{\BBOQ}\APACrefatitle {Visualizing Data Using T-{{SNE}}} {Visualizing data
  using t-{{SNE}}}.{\BBCQ}
\newblock
\APACjournalVolNumPages{Journal of Machine Learning
  Research}{9}{86}{2579--2605}.
\PrintBackRefs{\CurrentBib}

\bibitem [\protect \citeauthoryear {%
Vaswani%
\ \protect \BOthers {.}}{%
Vaswani%
\ \protect \BOthers {.}}{%
{\protect \APACyear {2017}}%
}]{%
VaswaniShazeerParmarEtAl17b}
\APACinsertmetastar {%
VaswaniShazeerParmarEtAl17b}%
\begin{APACrefauthors}%
Vaswani, A.%
, Shazeer, N.%
, Parmar, N.%
, Uszkoreit, J.%
, Jones, L.%
, Gomez, A\BPBI N.%
\BDBL {}Polosukhin, I.%
\end{APACrefauthors}%
\unskip\
\newblock
\APACrefYearMonthDay{2017}{}{}.
\newblock
{\BBOQ}\APACrefatitle {Attention Is All You Need} {Attention is all you
  need}.{\BBCQ}
\newblock
\BIn{} I.~Guyon\ \BOthers {.}\ (\BEDS), \APACrefbtitle {Adv. {{Neur}}. {{Inf}}.
  {{Proc}}. {{Sys}}. 30} {Adv. {{Neur}}. {{Inf}}. {{Proc}}. {{Sys}}. 30}\
  (\BPGS\ 5998--6008).
\newblock
\APACaddressPublisher{{long beach, CA, USA}}{}.
\PrintBackRefs{\CurrentBib}

\end{thebibliography}

\end{document}